\documentclass[10pt,twocolumn,letterpaper]{article}
\usepackage[pagenumbers]{cvpr}   % arXiv version: camera-ready layout with page numbers

\usepackage[utf8]{inputenc}
\usepackage[T1]{fontenc}
\usepackage{booktabs}
\usepackage{amsmath,amssymb}
\usepackage{graphicx}
\usepackage{multirow}
\usepackage{xcolor}
\usepackage{enumitem}
\usepackage{float}
\setlist{nosep,leftmargin=*}

\definecolor{cvprblue}{rgb}{0.21,0.49,0.74}
\usepackage[pagebackref,breaklinks,colorlinks,allcolors=cvprblue]{hyperref}

\newcommand{\dB}{\,dB}
\def\paperID{}
\def\confName{arXiv}
\def\confYear{2026}

\title{What Does LOL Measure? Hidden Label Shifts in Low-Light Enhancement Benchmarks}

\author{Adityan Thirumani Srinivasan\\
University of Wisconsin--Madison
}

\begin{document}
\maketitle

\begin{abstract}
LOL-v1 and LOLv2-Real are the standard benchmarks for supervised low-light image enhancement, and several leading methods ship hand-set test-time adjustments with their released evaluation scripts. We audit these benchmarks with 18 public checkpoints of eight methods (11 supervised, 7 trained without normal-light labels) under a single metric implementation that reproduces every published number to within 0.09\dB.
We find that the benchmarks' test labels differ from their training labels in ways the training data do not reveal.
(i)~The 15 LOL-v1 test labels are colour-processed differently from the training labels: the same 15 captures are distributed in LOLv2 with labels of identical luminance but $1.48\times$ lower saturation (range $1.27$--$1.61\times$), and LOL-v1's own training labels match the LOLv2 rendering. We confirm this on the official release.
(ii)~The LOLv2-Real test set is the contiguous block of images 690--789, whose labels are 14\% darker than the training labels (Cliff's $\delta=-0.45$, $p<10^{-12}$), with no comparable shift in the input statistics we examined; methods trained without any labels score up to 5.5\dB\ higher on test than on train.
(iii)~All 15 LOL-v1 test inputs reappear in LOLv2-Real (12 in its training set), extending the cross-benchmark overlap reported in concurrent work.
The released test-time settings are consistent with compensating for these shifts: the saturation boost shipped for LOL-v1 and the brightness reduction shipped for LOLv2-Real both point in the direction of the shift, and the reduction equals the test-set optimum of our grid to two decimals.
Separating what can be calibrated on training data from what only test-set tuning finds, the test-only residual has median 0.00\dB\ on LOLv2-Syn, which has no shift, but 0.37\dB\ on LOL-v1 and 0.92\dB\ on LOLv2-Real (up to 3.9\dB). Under the shift, test-set checkpoint selection inflates a from-scratch model by 0.9--2.0\dB\ on LOLv2-Real by selecting an under-trained, dark-output checkpoint, and removing the released settings changes which method ranks first. We give concrete reporting recommendations.
\end{abstract}

% =====================================================================
\section{Introduction}
\label{sec:intro}

Paired benchmarks make progress measurable only if the test set is a fair sample of the task the training set defines. In supervised low-light image enhancement (LLIE), three paired sets dominate reported results: LOL-v1 \citep{wei2018deep} with 485 training and 15 test pairs, and the real-captured and synthetic subsets of LOLv2 \citep{yang2021sparse} with 689/100 and 900/100 pairs. A new method is typically ranked by PSNR and SSIM on these test sets, often to two decimals.

Two practices make these numbers fragile. First, released evaluation scripts sometimes contain hand-set post-hoc adjustments that apply only at test time; for example, HVI-CIDNet \citep{yan2025hvi} multiplies output saturation by 1.3 on LOL-v1 and rescales brightness by 0.80--0.82 on LOLv2-Real, and PairLIE \citep{fu2023learning} lowers its illumination exponent from 0.2 to 0.14 ``for LOL''. Second, public training configurations commonly validate on the test split and keep the best-scoring checkpoint. Both practices are well known to bias estimates in general \citep{dwork2015reusable,roelofs2019meta,lipton2019troubling}; the question for LLIE is how much, and why the settings that are tuned help at all.

This paper answers that question by examining the benchmarks themselves, and finds that the effect of test-set tuning is concentrated where the test labels differ from the training labels. Throughout, we keep three failure modes apart, because they interact but are not the same: a \emph{label shift} between a benchmark's training and test labels (\S\ref{sec:lolv1}--\ref{sec:lolv2}), \emph{overlap} between benchmarks, which matters only for cross-benchmark training and evaluation (\S\ref{sec:overlap}), and \emph{adaptive tuning} on the test set, through post-hoc settings or checkpoint selection (\S\ref{sec:calib}--\ref{sec:selection}). Our contributions are:
\begin{enumerate}
  \item \textbf{A colour-processing shift in the LOL-v1 test labels} (\S\ref{sec:lolv1}). The 15 LOL-v1 test inputs are pixel-identical to 15 LOLv2-Real inputs, but their labels are $1.48\times$ more saturated than the LOLv2 labels of the same captures, with luminance unchanged to within 0.011. LOL-v1's training labels match the less saturated rendering. We verify this on the official download, not a mirror.
  \item \textbf{A contiguous-split label shift in LOLv2-Real} (\S\ref{sec:lolv2}). The test set is images 690--789, and its labels are significantly darker and less red than the training labels while the inputs are statistically indistinguishable. Seven checkpoints trained without labels score 2.4--5.5\dB\ \emph{higher} on test than on train.
  \item \textbf{A decomposition of test-time tuning} into a part recoverable from training data and a test-only residual, across 32 checkpoint--dataset pairs (\S\ref{sec:calib}). The residual vanishes on LOLv2-Syn, where there is no shift, and is largest on LOLv2-Real.
  \item \textbf{A demonstration of checkpoint selection under shift} (\S\ref{sec:selection}): in a controlled from-scratch model on LOLv2-Real, test-set selection picks a checkpoint from early training whose test score exceeds the validation-selected checkpoint by 0.9--2.0\dB, while its validation score is about 3\dB\ lower.
  \item \textbf{Consequences and recommendations} (\S\ref{sec:rank}, \S\ref{sec:recs}): removing released settings changes the first-ranked method on both real benchmarks, and Y-channel PSNR adds 1.9--3.7\dB, more than separates most methods.
\end{enumerate}
As supporting evidence we also report the cross-benchmark overlap (\S\ref{sec:overlap}); its main count was found independently and concurrently by \citet{pan2026leaky}, and Table~\ref{tab:novelty} separates what each work establishes. We do not claim that any method's architecture is ineffective, nor do we know how the released settings were chosen; the released checkpoints reproduce their papers' numbers, and the issue we document is what those numbers measure.

% =====================================================================
\section{Related work}
\label{sec:related}

\paragraph{Benchmark validity.}
Test-set reuse and adaptive analysis can inflate reported accuracy \citep{dwork2015reusable}, although large classification benchmarks have proven fairly robust to it \citep{roelofs2019meta}. Re-collected test sets expose distribution shift between a benchmark's construction and its use \citep{recht2019imagenet}, near-duplicates between training and test sets inflate scores \citep{barz2020we}, and evaluation-protocol choices can account for apparent progress \citep{musgrave2020metric,bouthillier2021accounting}. Leakage is a recognised source of irreproducibility across ML-based science \citep{kapoor2023leakage}. Our setting differs in that the shift lies in how the \emph{labels} were produced, and in that a hand-set post-processing step happens to compensate for it.

\paragraph{LLIE evaluation.}
LOL-v1 \citep{wei2018deep} and LOLv2 \citep{yang2021sparse} are the standard paired benchmarks; surveys document the variety of metrics in use \citep{li2022low}. The ``GT-mean'' protocol, which rescales each output to the mean intensity of its label before scoring, was used by KinD \citep{zhang2019kindling} and LLFlow \citep{wang2022low} and is reported alongside normal PSNR by recent methods; the Retinexformer authors explicitly advise against it because it uses the label at test time \citep{cai2023retinexformer}.
Concurrently with this work, \citet{pan2026leaky} report that 91 LOLv2-Real test labels are pixel-identical to LOL-v1 training labels, that 14 of 15 LOL-v1 test scenes have a training frame of the same scene and viewpoint, and that a nearest-neighbour lookup of training labels scores competitively on LOL-v1. Our overlap count agrees with theirs. Table~\ref{tab:novelty} compares the two works. The label-processing shift in LOL-v1, the contiguous-split label shift in LOLv2-Real, the evidence from label-free methods, the audit of released test-time settings, the calibration decomposition and the selection experiment are, to our knowledge, new.

\begin{table*}[t]
\centering
\caption{Scope of this work and the concurrent work of \citet{pan2026leaky}. \checkmark: established; (\checkmark): reported in passing; --: not addressed.}
\label{tab:novelty}
\small
\begin{tabular}{p{0.58\linewidth}cc}
\toprule
Finding or analysis & \citet{pan2026leaky} & Ours \\
\midrule
LOLv2-Real test labels identical to LOL-v1 training labels (91/100) & \checkmark & \checkmark \\
LOL-v1 test scenes with same-scene training twins & \checkmark (14/15) & (\checkmark) (5/15) \\
Nearest-neighbour lookup baseline; leakage-free split lists & \checkmark & -- \\
LOL-v1 test inputs inside LOLv2-Real, with re-rendered labels & -- & \checkmark \\
LOL-v1 test labels $1.48\times$ more saturated than the training rendering & -- & \checkmark \\
LOLv2-Real contiguous split with darker test labels & -- & \checkmark \\
Label-free methods as a control for label shift & -- & \checkmark \\
Audit of released test-time settings & -- & \checkmark \\
Train-fitted vs.\ test-fitted calibration decomposition (32 pairs) & -- & \checkmark \\
Controlled checkpoint-selection experiment & -- & \checkmark \\
Recommendation against test-set selection; GT-mean and brightness reporting & \checkmark & \checkmark \\
\bottomrule
\end{tabular}
\end{table*}

% =====================================================================
\section{Setup}
\label{sec:setup}

\paragraph{Data.}
We use LOL-v1 \citep{wei2018deep} (485/15) and the real and synthetic subsets of LOLv2 \citep{yang2021sparse} (689/100 and 900/100). Pixel-level comparisons use the decoded RGB array and its MD5 hash. Label-provenance results in \S\ref{sec:lolv1} are confirmed on the official LOL-v1 archive linked from the dataset's project page and on the official LOLv2 archive linked from its authors' repository; the widely used Kaggle mirror we ran most experiments on is pixel-identical to the official LOL-v1 release for all 15 test labels.

\paragraph{Checkpoints.}
We evaluate the released weights of three supervised methods, HVI-CIDNet \citep{yan2025hvi}, Retinexformer \citep{cai2023retinexformer} and IAT \citep{cui2022you}, on every benchmark for which a checkpoint exists (11 checkpoints), and seven checkpoints of five methods trained without normal-light labels, each evaluated on all three benchmarks: Zero-DCE \citep{guo2020zero}, Zero-DCE++ \citep{li2021learning}, SCI in its three released variants \citep{ma2022toward}, RUAS \citep{liu2021retinex} and PairLIE \citep{fu2023learning}. The latter group cannot have fitted LOL's label rendering, which makes it a control for shifts in the labels.

\paragraph{Metrics.}
PSNR and SSIM follow an exact re-implementation of HVI-CIDNet's \texttt{measure.py} (8-bit outputs, RGB PSNR, Gaussian-window SSIM averaged over channels); LPIPS uses AlexNet features \citep{zhang2018unreasonable}. GT-mean scores rescale the output to the label's mean grey level. Every number in this paper is computed by the same code.

\paragraph{Protocols.}
\emph{Native} denotes a checkpoint run through its authors' released evaluation script, including any test-time setting. \emph{Plain} denotes the same checkpoint with those settings removed. Methods without test-time settings score identically under both.

\paragraph{Post-hoc calibration.}
To ask what a test-time setting could achieve, we apply a global two-parameter adjustment to a model's output $y$ with luma $L(y)$:
$y' = \mathrm{clip}\big(b\cdot \mathrm{clip}(L + s\,(y-L))\big)$, with saturation $s\in\{0.60, 0.65,\dots,1.80\}$ and brightness $b\in\{0.60,0.62,\dots,1.40\}$. The \emph{oracle} setting maximises mean test PSNR; the \emph{honest} setting maximises mean PSNR on the training images and is then applied to test. The oracle gain minus the honest gain is the \emph{test-only residual}: improvement that is available only by looking at the test labels. The oracle is optimistic by construction, since it selects on the evaluation data; we use the residual as an operational measure of how far the test optimum departs from the training optimum, not as a causal estimate of inflation due to label shift alone.

\paragraph{Statistics.}
Train--test comparisons of image statistics use two-sided Mann--Whitney $U$ tests \citep{mann1947test} and Cliff's $\delta$ \citep{cliff1993dominance} ($|\delta|\ge 0.33$ medium, $\ge 0.47$ large). Confidence intervals use 10{,}000 bootstrap resamples of test images \citep{efron1994introduction}; ranking probabilities are the fraction of resamples in which a method ranks first. Colourfulness follows \citet{hasler2003measuring}. We test ten image statistics per benchmark and report Holm-adjusted $p$-values \citep{holm1979simple} where a conclusion depends on them.

All experiments ran on a single free-tier GPU (NVIDIA T4) or CPU.

% =====================================================================
\section{Reproduction and the released test-time settings}
\label{sec:repro}

\begin{table}[t]
\centering
\caption{Released checkpoints re-scored with one metric implementation. ``Paper'' is the number the authors report for the metric they report (PSNR unless marked GT-mean). All reproduce within 0.09\dB.}
\label{tab:repro}
\footnotesize
\setlength{\tabcolsep}{2.5pt}
\resizebox{\linewidth}{!}{\begin{tabular}{llllrrr}
\toprule
Method & Benchmark & Checkpoint & Metric & Paper & Ours & $\Delta$ \\
\midrule
HVI-CIDNet & LOL-v1 & w/o perc. & GT-mean & 28.14 & 28.23 & $+0.09$ \\
HVI-CIDNet & LOL-v1 & w/ perc. & GT-mean & 27.71 & 27.73 & $+0.02$ \\
Retinexformer & LOL-v1 & -- & PSNR & 25.16 & 25.15 & $-0.01$ \\
IAT & LOL-v1 & -- & PSNR & 23.38 & 23.38 & $+0.00$ \\
\midrule
HVI-CIDNet & LOLv2-Real & bestPSNR & PSNR & 23.90 & 23.89 & $-0.01$ \\
HVI-CIDNet & LOLv2-Real & bestSSIM & GT-mean & 28.39 & 28.39 & $-0.00$ \\
Retinexformer & LOLv2-Real & -- & PSNR & 22.80 & 22.79 & $-0.01$ \\
IAT & LOLv2-Real & -- & PSNR & 23.50 & 23.50 & $-0.00$ \\
\midrule
HVI-CIDNet & LOLv2-Syn & w/o perc. & GT-mean & 29.57 & 29.65 & $+0.08$ \\
HVI-CIDNet & LOLv2-Syn & w/ perc. & GT-mean & 29.37 & 29.39 & $+0.02$ \\
Retinexformer & LOLv2-Syn & -- & PSNR & 25.67 & 25.67 & $-0.00$ \\
\bottomrule
\end{tabular}}
\end{table}

Table~\ref{tab:repro} shows that every released checkpoint reproduces its paper's number to within 0.09\dB\ under our implementation, so the effects below are not implementation artefacts. Swapping between the PSNR implementations of HVI-CIDNet, Retinexformer, IAT and scikit-image changes RGB PSNR by at most 0.004\dB\ and SSIM by at most 0.003.

\begin{table*}[t]
\centering
\caption{Released test-time settings and what they are worth. ``Released'' is the native protocol; ``removed'' is the same checkpoint without the setting. The last column gives the optimal saturation ($s$) or brightness ($b$) of our post-hoc grid on training images and on test images, applied to the plain output.}
\label{tab:knobs}
\footnotesize
\setlength{\tabcolsep}{3.5pt}
\begin{tabular}{lllrrrrl}
\toprule
& & & \multicolumn{2}{c}{PSNR} & \multicolumn{2}{c}{GT-mean PSNR} & Grid optimum \\
\cmidrule(lr){4-5}\cmidrule(lr){6-7}
Checkpoint & Benchmark & Released setting & removed & released & removed & released & train / test \\
\midrule
CIDNet w/o perc. & LOL-v1 & saturation $\times1.3$ & 23.08 & 23.61 & 26.95 & 28.23 & $s$ 1.00 / 1.45 \\
CIDNet w/ perc.  & LOL-v1 & saturation $\times1.3$ & 23.22 & 23.84 & 26.65 & 27.73 & $s$ 1.00 / 1.50 \\
CIDNet bestPSNR  & LOLv2-Real & brightness $\alpha{=}0.80$ & 20.56 & 23.89 & 27.96 & 27.95 & $b$ 1.02 / 0.80 \\
CIDNet bestSSIM  & LOLv2-Real & brightness $\alpha{=}0.82$ & 20.78 & 23.89 & 28.39 & 28.39 & $b$ 1.02 / 0.82 \\
PairLIE & LOL-v1 & exponent 0.14 (vs.\ 0.2) & 18.50 & 19.52 & 23.51 & 23.16 & -- \\
PairLIE & LOLv2-Real & exponent 0.14 (vs.\ 0.2) & 19.88 & 18.76 & 24.03 & 23.64 & -- \\
PairLIE & LOLv2-Syn & exponent 0.14 (vs.\ 0.2) & 19.06 & 18.34 & 21.67 & 21.23 & -- \\
\bottomrule
\end{tabular}
\end{table*}

Table~\ref{tab:knobs} lists the test-time settings we found in released evaluation code. HVI-CIDNet's LOL-v1 saturation boost is worth $+1.08$ to $+1.28$\dB\ GT-mean PSNR, the metric its paper reports for LOL-v1; its LOLv2-Real brightness factor is worth $+3.1$ to $+3.3$\dB\ normal PSNR and nothing in GT-mean PSNR, which by construction ignores global brightness. The brightness factors released for LOLv2-Real, 0.80 and 0.82, are exactly the brightness components of the test-set optima of our grid, whereas the training-set optimum is 1.02. PairLIE's LOL exponent helps by 1.02\dB\ on LOL-v1 and hurts by 1.13\dB\ and 0.73\dB\ on the other two benchmarks. In addition, the public training configurations of HVI-CIDNet, Retinexformer and IAT validate on the test split, so their released checkpoints are also selected on test data.

The natural reading of Table~\ref{tab:knobs} is overfitting to the test set. The next three sections show something more specific: in each case the setting moves outputs in the direction of a measured difference between the test labels and the training labels. We observe this alignment; we do not know how the settings were originally chosen.

% =====================================================================
\section{LOL-v1: the test labels are colour-processed differently}
\label{sec:lolv1}

\begin{figure*}[t]
\centering
\includegraphics[width=\linewidth]{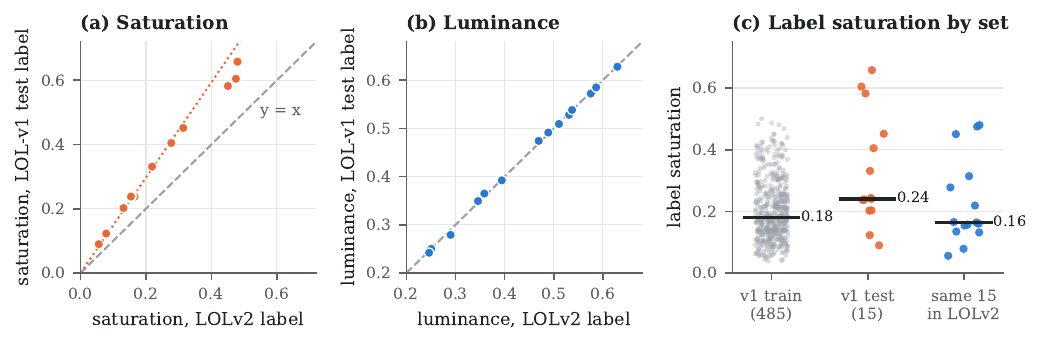}
\caption{The 15 LOL-v1 test inputs are pixel-identical to 15 LOLv2-Real inputs, but their labels differ. (a)~LOL-v1 test labels are more saturated than the LOLv2 labels of the same captures (dotted: median ratio, $1.48\times$; dashed: $y=x$). (b)~Luminance is unchanged. (c)~LOL-v1's training labels (grey) match the LOLv2 rendering (blue), not the test labels (orange). Bars are medians. Official releases.}
\label{fig:lolv1}
\end{figure*}

\paragraph{The same captures with two different labels.}
Every one of the 15 LOL-v1 test inputs has a pixel-identical copy in LOLv2-Real: 12 in its training set and 3 (images 748, 778, 780) in its test set. None of the corresponding labels is identical; the two versions of each label agree to only 24.6--39.0\dB\ PSNR (median 30.8\dB). Figure~\ref{fig:lolv1}(a,b) shows why. The LOL-v1 label is more saturated in all 15 cases, by a factor of 1.27--1.61 (median 1.48), while mean luminance differs by at most 0.011 on a $[0,1]$ scale. The direction is the same for all 15 images (two-sided sign test $p=6\times10^{-5}$), so this comparison does not rest on the 15-versus-485 test above. The pattern is consistent with a global saturation adjustment applied to one version and not the other.

\paragraph{The training labels follow the other rendering.}
LOL-v1's 485 training labels have median saturation 0.180, the LOLv2 versions of the 15 test labels have 0.164, and the LOL-v1 test labels have 0.240 (Figure~\ref{fig:lolv1}c). Among the ten image statistics we measured, only label saturation ($1.34\times$, $p=0.004$, Holm-adjusted $0.036$, $\delta=0.44$) and label colourfulness ($1.60\times$, $p=0.0003$, adjusted $0.003$, $\delta=0.55$) differ between LOL-v1's training and test sets; input luminance, input saturation, input noise, label luminance and the required gain do not ($p\ge0.47$, Table~\ref{tab:rawshift}). Several test labels also have a counterpart among the training labels: for five of them (665, 669, 748, 778, 780), the LOLv2 version matches a LOL-v1 training label (664, 668, 747, 775, 779) in both mean luminance and saturation to four decimals, consistent with the same-scene twins reported by \citet{pan2026leaky}. A model trained on LOL-v1 therefore learns one colour rendering and is scored against another.

\paragraph{Provenance.}
Because a re-encoded mirror could introduce such a difference, we repeated the comparison on the official LOL-v1 archive linked from the dataset's project page and the official LOLv2 archive linked from the authors' repository. The official LOL-v1 test labels are pixel-identical to the mirror's for all 15 images, and the official LOLv2 labels reproduce the lower saturation. We cannot tell from the data which rendering is the ``original''; the finding is that the benchmark's train and test labels differ in rendering.

\paragraph{Every model is undersaturated on test, including models that never saw a label.}
The four supervised LOL-v1 checkpoints match label saturation on their training images (output-to-label ratio 0.94--0.98) but fall to 0.59--0.71 on test (Appendix Table~\ref{tab:drift}). The label-free checkpoints show the same drop: six of them oversaturate relative to the labels on both splits, with the ratio falling from 2.4--2.8 on train to 1.4--1.6 on test, and PairLIE falls from 1.46 to 0.84. Since their outputs depend only on inputs, and input saturation does not differ between splits ($p=0.62$), the drop can only come from the labels. The saturation that maximises test PSNR for the supervised checkpoints is $s=1.30$--$1.50$, against $1.00$ on training images, and HVI-CIDNet's released $\times1.3$ boost moves its outputs most of the way towards the test labels' rendering, consistent with compensating for the shift.

% =====================================================================
\section{LOLv2-Real: a contiguous split with darker labels}
\label{sec:lolv2}

\begin{figure*}[t]
\centering
\includegraphics[width=\linewidth]{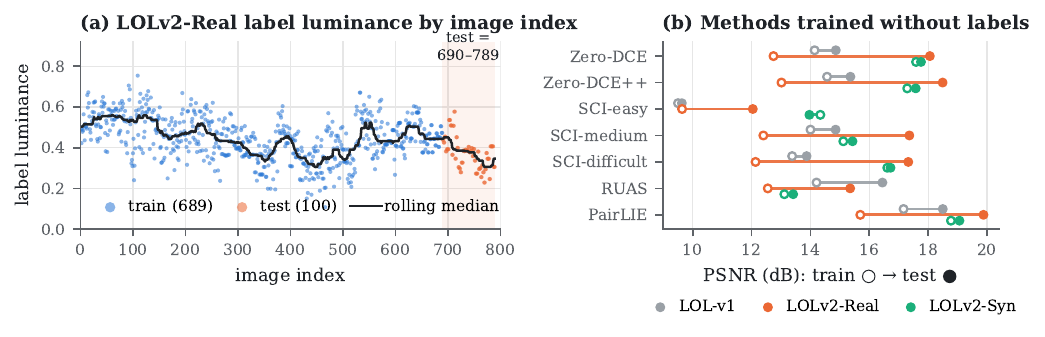}
\caption{(a)~LOLv2-Real's test set (shaded) is the last 100 images by index; their labels are darker than the training labels (black: rolling median over 41 images). (b)~Seven checkpoints trained without normal-light labels, scored on each benchmark's training (open) and test (filled) images. On LOLv2-Real (orange) every one scores higher on test, by up to 5.5\dB; on LOLv2-Syn (green) train and test agree.}
\label{fig:lolv2}
\end{figure*}

\begin{table*}[t]
\centering
\caption{Train versus test image statistics. Ratio of medians (test/train), two-sided Mann--Whitney $p$ (unadjusted), and Cliff's $\delta$ (test vs.\ train). Bold: medium or large effect ($|\delta|\ge0.33$). The label statistics shift; the input statistics we examined do not shift comparably.}
\label{tab:rawshift}
\small
\setlength{\tabcolsep}{4pt}
\begin{tabular}{lrrrrrrrrr}
\toprule
& \multicolumn{3}{c}{LOL-v1 (485 / 15)} & \multicolumn{3}{c}{LOLv2-Real (689 / 100)} & \multicolumn{3}{c}{LOLv2-Syn (900 / 100)} \\
\cmidrule(lr){2-4}\cmidrule(lr){5-7}\cmidrule(lr){8-10}
Statistic & ratio & $p$ & $\delta$ & ratio & $p$ & $\delta$ & ratio & $p$ & $\delta$ \\
\midrule
Input luminance & 1.15 & 0.54 & 0.09 & 0.84 & 0.29 & $-0.07$ & 0.85 & 0.048 & $-0.12$ \\
Input saturation & 0.94 & 0.62 & $-0.08$ & 0.93 & 0.012 & $-0.16$ & 1.02 & 0.98 & 0.00 \\
Input noise (MAD of Laplacian) & 1.02 & 0.53 & 0.10 & 1.11 & 0.22 & 0.08 & 0.95 & 0.53 & $-0.04$ \\
Required gain (label/input lum.) & 0.84 & 0.48 & $-0.11$ & 0.90 & 0.039 & $-0.13$ & 1.07 & 0.09 & 0.10 \\
\midrule
Label luminance & 1.08 & 0.95 & 0.01 & 0.86 & $3\!\times\!10^{-13}$ & $\mathbf{-0.45}$ & 0.96 & 0.18 & $-0.08$ \\
Label red channel & 1.09 & 0.73 & 0.05 & 0.79 & $3\!\times\!10^{-18}$ & $\mathbf{-0.54}$ & 0.93 & 0.058 & $-0.12$ \\
Label saturation & 1.34 & 0.004 & $\mathbf{0.44}$ & 0.84 & 0.037 & $-0.13$ & 1.04 & 0.53 & 0.04 \\
Label colourfulness & 1.60 & 0.0003 & $\mathbf{0.55}$ & 0.79 & $2\!\times\!10^{-5}$ & $-0.27$ & 1.05 & 0.35 & 0.06 \\
\bottomrule
\end{tabular}
\end{table*}

\paragraph{The split.}
LOLv2-Real's training set is images 1--689 and its test set images 690--789 (Figure~\ref{fig:lolv2}a). This is not a random split, and the last block differs: its labels are 14\% darker than the training labels ($p=3\times10^{-13}$, $\delta=-0.45$), with the largest change in the red channel ($0.79\times$, $\delta=-0.54$), and are less colourful (Table~\ref{tab:rawshift}). We find no shift of comparable size in the input statistics we examined (input luminance $p=0.29$, $\delta=-0.07$; input saturation, noise and required gain all $|\delta|\le0.16$, none significant after Holm adjustment), whereas label luminance, all three label channels and label colourfulness remain significant after adjustment ($p_{\mathrm{Holm}}\le10^{-4}$). Matching summary statistics do not prove the input distributions identical; scene content or difficulty could differ in ways these summaries miss, and we return to this in \S\ref{sec:limits}. The same block of indices appears in LOL-v1's numbering: 85 of the 100 test pairs (input and label) occur in LOL-v1's training set under the same file index (\S\ref{sec:overlap}). Within LOL-v1's training set, the images that became LOLv2-Real's test set are also darker than the rest (median label luminance 0.38 vs.\ 0.48, $p=4\times10^{-18}$), which shows that the shift comes from which scenes were assigned to test, not from re-processing.

\paragraph{Label-free methods score far higher on test.}
If the test labels are darker and current methods tend to under-brighten, the test set should reward dark outputs. All seven label-free checkpoints score higher on LOLv2-Real's test set than on its training set: Zero-DCE by 5.3\dB, Zero-DCE++ by 5.5\dB, SCI by 2.4--5.2\dB, RUAS by 2.8\dB\ and PairLIE by 4.2\dB\ (Figure~\ref{fig:lolv2}b). On LOLv2-Syn, which has no shift, the same checkpoints change by $-0.4$ to $+0.3$\dB. Their output-to-label brightness ratio rises from train to test for all seven (e.g.\ Zero-DCE 0.61 to 0.73). The supervised checkpoints show the same pattern: IAT's released LOLv2 checkpoint scores 18.2\dB\ on training images and 23.5\dB\ on test, and Retinexformer 21.5\dB\ and 22.8\dB. HVI-CIDNet is too dark on its training images (ratio 0.94--1.01) and too bright on test (1.26), and its released brightness factor of 0.80 moves it to the test labels' brightness.

\paragraph{LOLv2-Syn as a control.}
On the synthetic subset, no label statistic differs between splits by more than a small effect ($|\delta|\le0.12$), every checkpoint's train and test output ratios agree, and no released method uses a test-time setting there.

% =====================================================================
\section{Overlap between the benchmarks}
\label{sec:overlap}

\begin{table*}[t]
\centering
\caption{Pixel-identical images between LOL-v1 and LOLv2-Real (MD5 of decoded RGB). Label counts are per image; some labels are shared by several inputs within a benchmark (LOL-v1: 500 label files, 414 distinct; LOLv2-Real: 789, 642). No label is shared between the training and test sets of the same benchmark.}
\label{tab:overlap}
\small
\begin{tabular}{llrl}
\toprule
Set & Found in & Identical & Note \\
\midrule
LOLv2-Real test labels (100) & LOL-v1 train & 91 & \citet{pan2026leaky} report 91 \\
LOLv2-Real test inputs (100) & LOL-v1 train & 86 & \\
LOLv2-Real train inputs (689) & LOL-v1 train & 399 & \\
LOL-v1 test inputs (15) & LOLv2-Real & 15 & 12 in LOLv2 train, 3 in LOLv2 test \\
LOL-v1 test labels (15) & LOLv2-Real & 0 & same captures, re-rendered labels (\S\ref{sec:lolv1}) \\
\bottomrule
\end{tabular}
\end{table*}

Table~\ref{tab:overlap} summarises the overlap. Any model trained on LOL-v1 and scored on LOLv2-Real has seen 91 of the 100 test labels, which matters for cross-dataset tables and for methods trained on the union of both sets; \citet{pan2026leaky} reach the same count and propose leakage-free subsets. The overlap also runs in the other direction: a model trained on LOLv2-Real has seen 12 of the 15 LOL-v1 test inputs, paired with the less saturated rendering of their labels. Combining the two training sets, which some methods do, therefore teaches both renderings of the same scenes.

% =====================================================================
\section{What test-set tuning recovers}
\label{sec:calib}

\begin{figure*}[t]
\centering
\includegraphics[width=\linewidth]{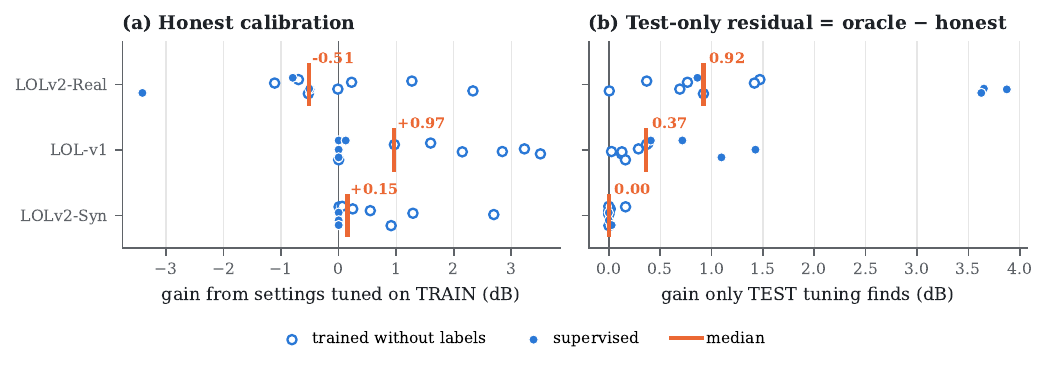}
\caption{Post-hoc saturation/brightness calibration for 32 checkpoint--benchmark pairs. (a)~Gain on test from the setting that is best on training images. (b)~Test-only residual: the oracle gain (best setting on test) minus the gain in (a). The residual is zero without a shift (LOLv2-Syn) and largest under the darker-label shift (LOLv2-Real).}
\label{fig:calib}
\end{figure*}

If a test-time setting fixes a systematic bias of the model, the same bias is visible on training images and a setting fitted there should transfer. If it fixes a train--test shift in the labels, it cannot. Figure~\ref{fig:calib} separates the two for all 32 checkpoint--benchmark pairs.

On LOLv2-Syn the training-fitted setting comes within 0.03\dB\ of the oracle gain for 9 of 10 checkpoints and the median residual is 0.00\dB. On LOL-v1, label-free methods have large, consistent brightness errors that a training-fitted setting fixes (gains up to 3.5\dB), leaving a residual of 0.03--0.38\dB; for the supervised checkpoints, which are already calibrated on training data, the training-fitted setting is the identity or close to it, and 0.41--1.43\dB\ of the oracle gain is test-only. On LOLv2-Real the training-fitted setting \emph{lowers} test PSNR for 8 of 11 checkpoints (by up to 3.4\dB\ for IAT), because the training labels are brighter than the test labels; the median residual is 0.92\dB\ and reaches 3.9\dB\ for HVI-CIDNet. In short, where a shift exists, most of the gain from test-time settings is gain that cannot be obtained from training data.

% =====================================================================
\section{Checkpoint selection under shift}
\label{sec:selection}

\begin{figure*}[t]
\centering
\includegraphics[width=\linewidth]{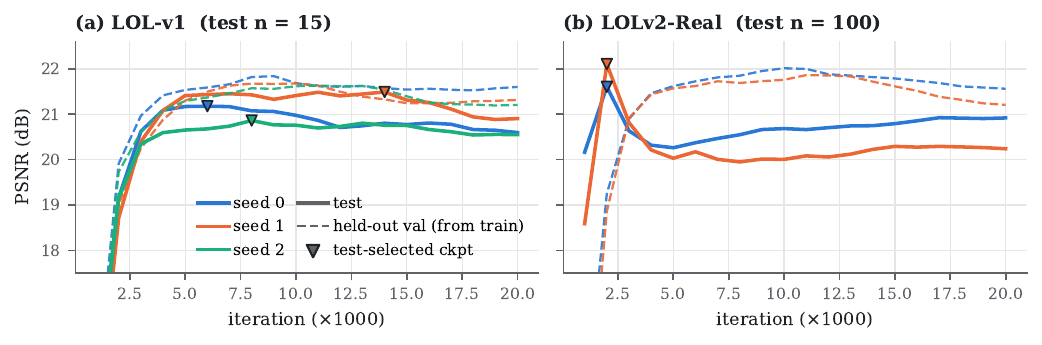}
\caption{A U-Net trained from scratch, scored every 1{,}000 iterations on a validation split held out from training (dashed) and on the test set (solid); triangles mark the checkpoint test-set selection would keep. On LOLv2-Real, test PSNR peaks at 2{,}000 iterations, when outputs are still dark, while validation PSNR keeps rising.}
\label{fig:selection}
\end{figure*}

\begin{table*}[t]
\centering
\caption{Selection inflation (test PSNR of the test-selected checkpoint minus that of the validation-selected checkpoint) and honest calibration (post-hoc setting fitted on the validation split) for five from-scratch runs.}
\label{tab:selection}
\small
\begin{tabular}{lcccccc}
\toprule
& & \multicolumn{2}{c}{Selected iteration} & & \multicolumn{2}{c}{Calibration gain on test (dB)} \\
\cmidrule(lr){3-4}\cmidrule(lr){6-7}
Benchmark & Seed & by val & by test & Inflation (dB) & honest (val-fitted) & oracle \\
\midrule
LOL-v1 & 0 & 9k & 6k & 0.12 & 0.10 & 0.58 \\
LOL-v1 & 1 & 10k & 14k & 0.08 & 0.10 & 0.68 \\
LOL-v1 & 2 & 11k & 8k & 0.16 & 0.09 & 0.69 \\
LOLv2-Real & 0 & 10k & 2k & 0.92 & 0.00 & 1.53 \\
LOLv2-Real & 1 & 11k & 2k & 2.03 & 0.39 & 2.21 \\
\bottomrule
\end{tabular}
\end{table*}

Released checkpoints are selected on the test split, so we measure what that selection is worth on a model we control. This experiment demonstrates a failure mechanism; with one architecture and five runs it does not estimate how often or how severely the mechanism affects other architectures and training recipes. We train a 0.88M-parameter U-Net (GroupNorm, residual output) from scratch for 20{,}000 iterations with L1 + $0.2\,(1-\mathrm{SSIM})$ loss, AdamW (learning rate $2\times10^{-4}$, one-cycle schedule), batch 8, 256-pixel crops and an EMA of weights, holding out 50 training pairs as a validation split. Every 1{,}000 iterations the EMA model is scored on validation and test.

On LOL-v1, validation and test curves move together and test-set selection adds 0.08--0.16\dB\ over validation selection (Table~\ref{tab:selection}). On LOLv2-Real it adds 0.92 and 2.03\dB, and in both runs it selects the checkpoint at 2{,}000 iterations (Figure~\ref{fig:selection}), whose validation PSNR (19.2 and 18.9\dB) is about 3\dB\ below the best validation checkpoint. The early checkpoint is still under-exposing, and the darker test labels reward it. The larger inflation on the 100-image test set than on the 15-image one indicates that, in this experiment, the shift matters more than test-set size.

Fitting the post-hoc setting on the held-out validation split recovers 14--16\% of the oracle calibration gain on LOL-v1 and 0--18\% on LOLv2-Real. Holding out validation data from training is necessary, but it does not recover a shift the training distribution does not contain; the test-only part has to be reported as such or removed, not tuned away.

% =====================================================================
\section{Consequences for rankings and metrics}
\label{sec:rank}

\begin{figure*}[t]
\centering
\includegraphics[width=\linewidth]{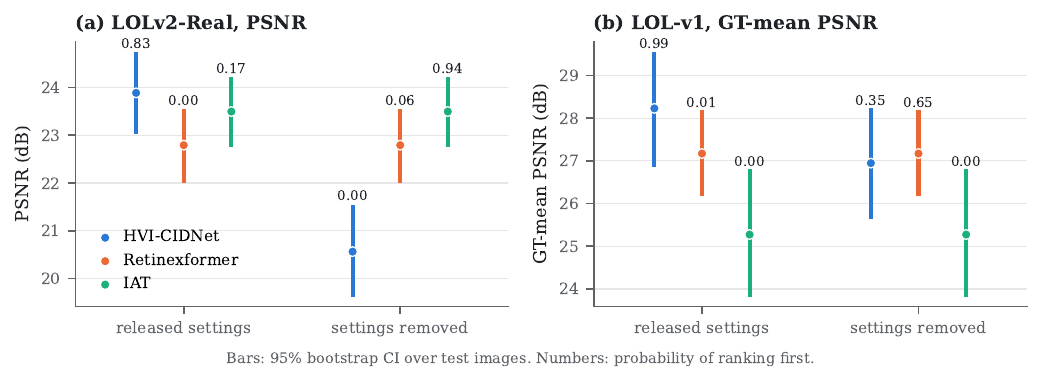}
\caption{Means and 95\% bootstrap intervals with and without the released test-time settings; numbers above bars give the probability of ranking first over 10{,}000 resamples. (a)~On LOLv2-Real, removing HVI-CIDNet's brightness setting moves the first rank to IAT. (b)~On LOL-v1, removing the saturation setting moves first rank in GT-mean PSNR to Retinexformer, with wide intervals over 15 images.}
\label{fig:rank}
\end{figure*}

\paragraph{Rankings.}
With released settings, HVI-CIDNet ranks first on LOLv2-Real PSNR with probability 0.83; without them its PSNR falls from 23.89 to 20.56\dB\ and IAT ranks first with probability 0.94 (Figure~\ref{fig:rank}a). On LOL-v1 GT-mean PSNR, HVI-CIDNet ranks first with probability 0.99 with its saturation setting and 0.35 without, behind Retinexformer at 0.65 (Figure~\ref{fig:rank}b). On LOL-v1 normal PSNR, Retinexformer ranks first under both protocols. With 15 test images, the 95\% interval for a single method's LOL-v1 PSNR spans up to 5.2\dB\ (HVI-CIDNet: 20.9--26.2\dB), wider than the spread between methods, so LOL-v1 rankings in either protocol should be read as uncertain.

\paragraph{Metric choices.}
GT-mean rescaling removes global brightness error, so it hides exactly the LOLv2-Real shift: HVI-CIDNet's GT-mean PSNR on LOLv2-Real is unchanged by its brightness setting (27.96 vs.\ 27.95\dB) while its normal PSNR moves by 3.3\dB. Computing PSNR on the luma channel instead of RGB raises scores by 1.9--3.7\dB\ for the same outputs, and the GT-mean implementations of HVI-CIDNet and Retinexformer differ by up to 0.13\dB. A paper that does not state its metric variant cannot be compared to one that does.

% =====================================================================
\section{Recommendations}
\label{sec:recs}

\begin{enumerate}
  \item \textbf{Report results without test-time settings}, or report both, and state any setting's value and how it was chosen. A setting is legitimate only if it is fitted on training or held-out training data.
  \item \textbf{Select checkpoints on a validation split held out from training}, and report the selection rule. Never select on the test split.
  \item \textbf{Treat LOL-v1 and LOLv2-Real test scores as measuring transfer to a differently rendered or differently exposed label set}, not in-distribution accuracy. Where possible, also report on LOLv2-Syn or another set without a known shift.
  \item \textbf{Do not train on LOL-v1 and evaluate on LOLv2-Real}, or vice versa, without removing the overlapping images; within-benchmark training and evaluation is not affected by this overlap (Table~\ref{tab:overlap}; split lists in our release and in \citealp{pan2026leaky}).
  \item \textbf{State the metric variant} (RGB or luma PSNR, SSIM window, GT-mean or not) and report bootstrap intervals; on LOL-v1, differences below about 1\dB\ are not resolved by 15 images.
  \item \textbf{Run label-free reference methods} on both splits of any new benchmark: a large train--test gap for a method that cannot fit the labels is a cheap test for label shift.
\end{enumerate}

% =====================================================================
\section{Limitations}
\label{sec:limits}

We audit the two LOL benchmarks and 18 released checkpoints of eight methods; other paired benchmarks (e.g.\ SID, SDSD, LSRW) may or may not share these issues. We establish that LOL-v1's training and test labels differ in rendering, but we cannot determine which processing step produced the difference. Our post-hoc calibration has two global parameters; richer calibrations could recover more from training data, though they cannot recover a shift absent from it. The selection experiment uses one architecture and five runs and demonstrates a mechanism rather than measuring its prevalence. The LOLv2-Real evidence that the shift lies in the labels rests on input summary statistics and on label-free controls; neither excludes every difference in scene content or difficulty between the two blocks of images. Our observations show that the released settings align with the shifts, not how those settings were chosen. Finally, the label-free control assumes those checkpoints were not tuned on LOL labels; their training data differ, but none uses LOL normal-light labels as targets.

\section*{Broader impact}
This work concerns measurement practice in image enhancement and has no direct societal risk. Its intended effect is to make reported comparisons more reliable; we have tried to describe released settings neutrally, since they reflect common practice rather than individual misconduct.

\section*{Reproducibility}
All analyses use public datasets and public checkpoints. Every CSV used for the tables and figures, and the script that draws the figures, are included with this paper's source; the full audit code (re-scoring, calibration grid, raw-data tests, overlap and provenance checks, from-scratch selection runs) will be released publicly. Every experiment ran on one NVIDIA T4 or on CPU; each from-scratch run takes about 65 minutes.

{\small
\bibliographystyle{ieeenat_fullname}
\bibliography{references}
}

% =====================================================================
\clearpage
\onecolumn
\appendix
\section{Per-checkpoint train/test behaviour}
\label{app:drift}

\begin{table}[H]
\centering
\caption{Supervised checkpoints, plain protocol. Ratios are output-to-label mean saturation and mean luminance, averaged over images. Train PSNR uses all training images.}
\label{tab:drift}
\small
\begin{tabular}{lllrrrrrr}
\toprule
& & & \multicolumn{2}{c}{PSNR} & \multicolumn{2}{c}{Saturation ratio} & \multicolumn{2}{c}{Brightness ratio} \\
\cmidrule(lr){4-5}\cmidrule(lr){6-7}\cmidrule(lr){8-9}
Method & Benchmark & Ckpt & train & test & train & test & train & test \\
\midrule
HVI-CIDNet & LOL-v1 & w/o perc. & 31.90 & 23.08 & 0.98 & 0.68 & 1.00 & 1.05 \\
HVI-CIDNet & LOL-v1 & w/ perc. & 28.06 & 23.22 & 0.94 & 0.64 & 1.00 & 1.07 \\
Retinexformer & LOL-v1 & -- & 27.07 & 25.15 & 0.98 & 0.71 & 1.00 & 0.97 \\
IAT & LOL-v1 & -- & 25.80 & 23.38 & 0.97 & 0.59 & 0.98 & 0.99 \\
\midrule
HVI-CIDNet & LOLv2-Real & bestPSNR & 22.47 & 20.56 & 0.87 & 0.68 & 0.94 & 1.26 \\
HVI-CIDNet & LOLv2-Real & bestSSIM & 23.73 & 20.78 & 0.94 & 0.77 & 1.01 & 1.26 \\
Retinexformer & LOLv2-Real & -- & 21.51 & 22.79 & 1.05 & 0.99 & 0.90 & 1.05 \\
IAT & LOLv2-Real & -- & 18.18 & 23.50 & 0.87 & 0.84 & 0.84 & 1.01 \\
\midrule
HVI-CIDNet & LOLv2-Syn & w/o perc. & 35.18 & 25.72 & 0.98 & 0.98 & 0.99 & 1.02 \\
HVI-CIDNet & LOLv2-Syn & w/ perc. & 32.74 & 25.16 & 0.97 & 0.97 & 1.00 & 1.01 \\
Retinexformer & LOLv2-Syn & -- & 28.73 & 25.67 & 0.96 & 0.95 & 1.00 & 1.02 \\
\bottomrule
\end{tabular}
\end{table}

\begin{table}[H]
\centering
\caption{Checkpoints trained without normal-light labels. Train and test PSNR and output-to-label ratios on each benchmark.}
\label{tab:labelfree}
\small
\setlength{\tabcolsep}{4pt}
\begin{tabular}{lrrrrrrrrr}
\toprule
& \multicolumn{3}{c}{LOL-v1} & \multicolumn{3}{c}{LOLv2-Real} & \multicolumn{3}{c}{LOLv2-Syn} \\
\cmidrule(lr){2-4}\cmidrule(lr){5-7}\cmidrule(lr){8-10}
Checkpoint & train & test & sat.\ ratio & train & test & bright.\ ratio & train & test & bright.\ ratio \\
& PSNR & PSNR & tr$\to$te & PSNR & PSNR & tr$\to$te & PSNR & PSNR & tr$\to$te \\
\midrule
Zero-DCE & 14.14 & 14.86 & 2.48$\to$1.40 & 12.74 & 18.06 & 0.61$\to$0.73 & 17.59 & 17.76 & 0.86$\to$0.83 \\
Zero-DCE++ & 14.56 & 15.36 & 2.49$\to$1.44 & 13.01 & 18.49 & 0.65$\to$0.75 & 17.29 & 17.58 & 0.88$\to$0.86 \\
SCI-easy & 9.50 & 9.63 & 2.65$\to$1.51 & 9.63 & 12.04 & 0.31$\to$0.32 & 14.34 & 13.96 & 0.73$\to$0.68 \\
SCI-medium & 14.00 & 14.85 & 2.61$\to$1.49 & 12.40 & 17.36 & 0.64$\to$0.70 & 15.12 & 15.43 & 1.06$\to$1.02 \\
SCI-difficult & 13.38 & 13.87 & 2.42$\to$1.36 & 12.13 & 17.33 & 0.58$\to$0.64 & 16.61 & 16.72 & 0.92$\to$0.89 \\
RUAS & 14.20 & 16.45 & 2.80$\to$1.62 & 12.55 & 15.35 & 0.89$\to$0.94 & 13.11 & 13.41 & 1.25$\to$1.23 \\
PairLIE & 17.16 & 18.50 & 1.46$\to$0.84 & 15.69 & 19.88 & 0.90$\to$1.03 & 18.78 & 19.06 & 1.08$\to$1.07 \\
\bottomrule
\end{tabular}
\end{table}

\section{Mechanism within the test set}
On LOLv2-Real, the per-image saturation ratio of the supervised checkpoints correlates positively with input luminance (Spearman $\rho=0.22$--$0.50$, $p\le0.03$): darker inputs are rendered less saturated. On LOL-v1 none of these correlations is significant with 15 images. We report this for completeness; the label shifts in \S\ref{sec:lolv1}--\ref{sec:lolv2} do not depend on it.

\end{document}